\documentclass[runningheads]{llncs}

\usepackage[mobile]{eccv}

\usepackage{eccvabbrv}
\usepackage{float}
\usepackage{graphicx}
\usepackage{booktabs}

\usepackage[accsupp]{axessibility}  
\renewcommand{\paragraph}[1]{\vspace{1.25mm}\noindent\textbf{#1}}
\usepackage{multicol, blindtext, multirow}
\newcommand{\Rows}[1]{\multirow{2}{*}{#1}}

\usepackage[pagebackref,breaklinks,colorlinks]{hyperref}

\usepackage{orcidlink}

\begin{document}

\title{VFusion3D: Learning Scalable 3D Generative Models from Video Diffusion Models}

\titlerunning{VFusion3D}

\author{Junlin Han\inst{1,2}* \and
Filippos Kokkinos\inst{1}* \and
Philip Torr\inst{2}}
\authorrunning{Junlin Han et al.}

\institute{GenAI, Meta \and
Torr Vision Group, University of Oxford\\
* Equal contribution\\
\email{junlinhan@meta.com, fkokkinos@meta.com, philip.torr@eng.ox.ac.uk }\\
Project page: \textcolor{red}{\href{https://junlinhan.github.io/projects/vfusion3d.html}{https://junlinhan.github.io/projects/vfusion3d.html}}}

\maketitle

\vspace{-10pt}
 \begin{figure}[htb]
     \centering
     \includegraphics[width = \textwidth]
     {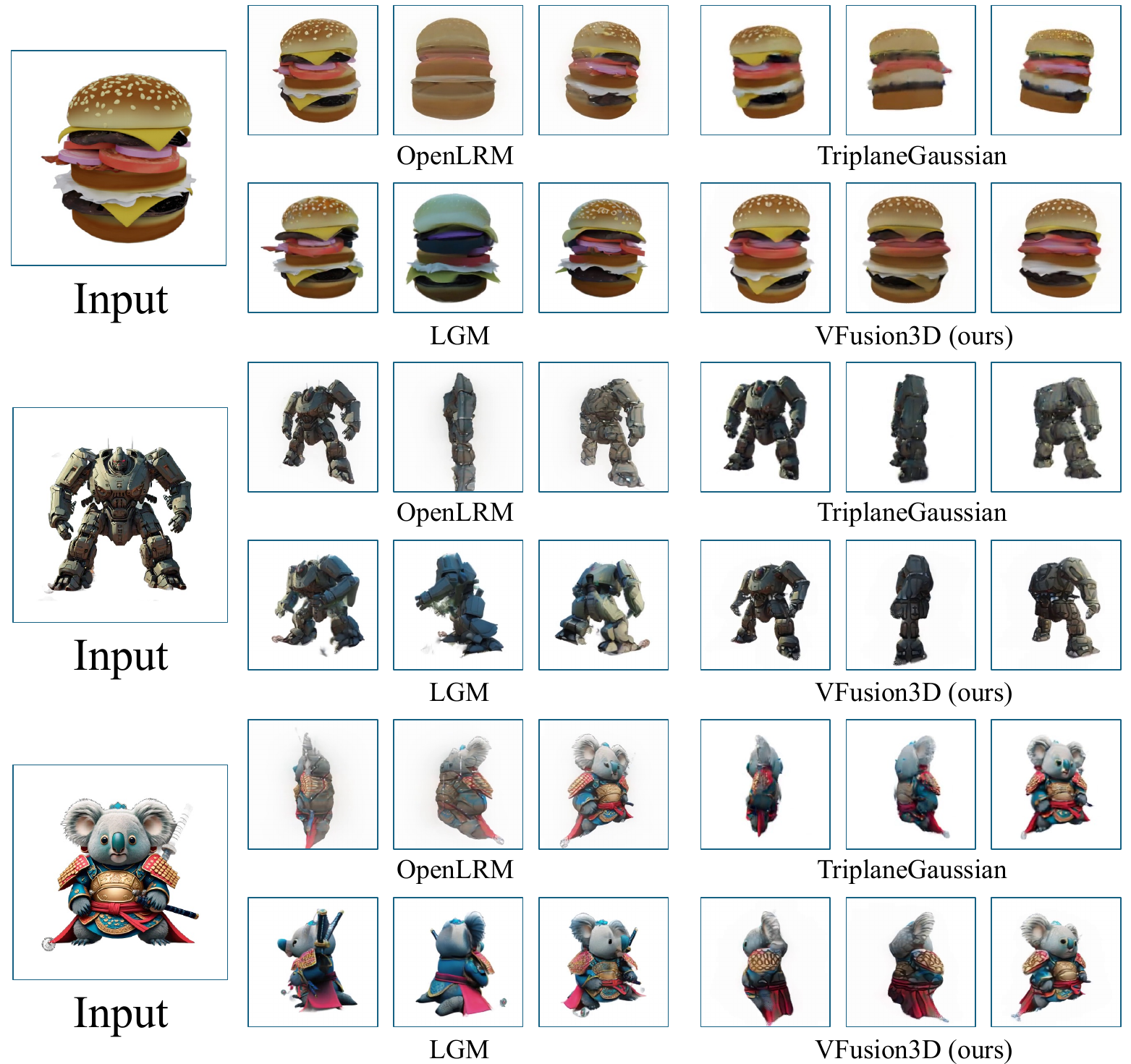}
        \caption{\textbf{Comparisons of large 3D reconstruction models}. Our VFusion3D reconstructs high-quality and 3D-consistent assets from a single input image.}
     \label{fig:fst_page}
\end{figure}
\vspace{-20pt}

\begin{abstract}
This paper presents a novel method for building scalable 3D generative models utilizing pre-trained video diffusion models. The primary obstacle in developing foundation 3D generative models is the limited availability of 3D data. Unlike images, texts, or videos, 3D data are not readily accessible and are difficult to acquire. This results in a significant disparity in scale compared to the vast quantities of other types of data.
To address this issue, we propose using a video diffusion model, trained with extensive volumes of text, images, and videos, as a knowledge source for 3D data. By unlocking its multi-view generative capabilities through fine-tuning, we generate a large-scale synthetic multi-view dataset to train a feed-forward 3D generative model. The proposed model, VFusion3D, trained on nearly 3M synthetic multi-view data, can generate a 3D asset from a single image in seconds and achieves superior performance when compared to current SOTA feed-forward 3D generative models, with users preferring our results over $90\%$ of the time.

\end{abstract}

\section{Introduction}
\label{sec:intro}
The advent of 3D datasets~\cite{chang2015shapenet,deitke2023objaverse, deitke2024objaversexl} and the development of advanced neural rendering methods~\cite{mescheder2019occupancy, park2019deepsdf, mildenhall2021nerf,kerbl233d-gaussian} have created new possibilities in computer vision and computer graphics. AI for 3D content creation has shown great potential in various fields, such as AR/VR/MR~\cite{li2022rt}, 3D gaming~\cite{hao2021gancraft,sun20233d}, and animation~\cite{kolotouros2024dreamhuman}.
As a result, there is a growing demand for the development of foundation 3D generative models that can efficiently create various high-quality 3D assets. Despite the clear need and potential, current methods fall short of expectations due to their production of low-quality textures and geometric issues such as floaters.

The primary obstacle in constructing foundation 3D generative models is the scarcity of 3D data available. Acquiring 3D data is a complex and challenging task. Unlike images, which can be easily captured using standard cameras, 3D data requires specialized equipment and capturing techniques. Another option is 3D modeling, a tedious process that can take up to weeks for a single asset. This complexity results in a pool of publicly available 3D assets that is limited, with the largest ones containing up to 10 million assets~\cite{deitke2024objaversexl,yu2023mvimgnet,chang2015shapenet}. In reality, only a small portion of them is usable since a decent percentage of 3D assets are duplicates and many of them lack texture or even a high-quality surface~\cite{shi2023mvdream, hong2023lrm, li2023instant3d}. 

Conversely, foundation models such as GPT~\cite{achiam2023gpt,touvron2023llama,team2023gemini} and Diffusion models~\cite{rombach2022high,dai2023emu,videoworldsimulators2024} have consistently showcased remarkable capabilities. These models have proven their prowess in handling intricate tasks, exhibiting advanced problem-solving skills, and delivering exceptional performance across a spectrum of challenges.
Such general and strong capabilities are derived from scaling on both data and model size~\cite{touvron2023llama,peebles2023scalable,videoworldsimulators2024}. The primary prerequisite for constructing foundation models is the accessibility of an extensive amount of high-quality training data, often surpassing billions in size.
This presents a clear discrepancy between the current state of 3D datasets and the requirements for training foundation 3D generative models. 

 \begin{figure}[t]
     \centering
     \includegraphics[width = \textwidth]
     {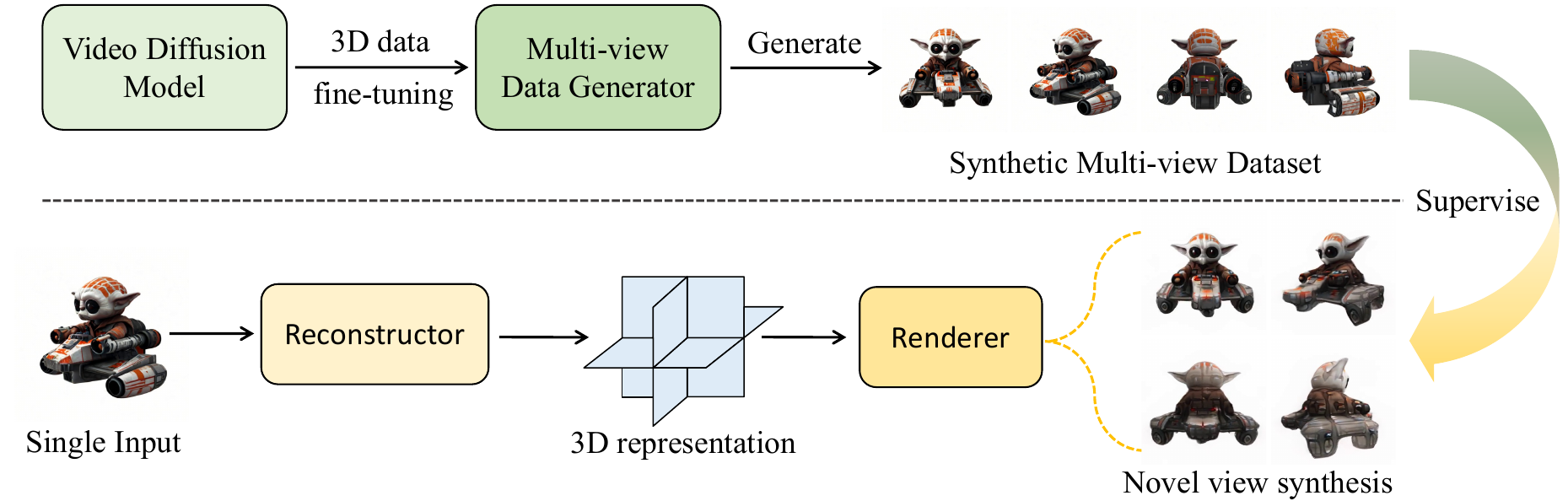}
    \caption{\textbf{The pipeline of VFusion3D}. We first use a small amount of 3D data to fine-tune a video diffusion model, transforming it into a multi-view vodeo generator that functions as a data engine. By generating a large amount of synthetic data, we train VFusion3D to generate a 3D representation and render novel views.
     }
     \label{fig:overall}
\end{figure}

In this paper, we propose to use of video diffusion models, trained on a large amount of texts, images, and videos, as a 3D data generator. Specifically, we use EMU Video~\cite{girdhar2023emuvideo} which has been trained with a variety of videos, including those with intricate camera movements and drone footage, and inherently contain cues about the 3D world. This suggests that video diffusion models have a latent understanding of how to generate videos with 3D consistency~\cite{melas2024im3d}.
By fine-tuning this model using rendered multi-view videos from 100K 3D data, we unlock the model's inherent capability to generate diverse, 3D-consistent, multi-view videos from text and image prompts. The resulting video diffusion model allows us to scale the 3D data necessary for learning foundation 3D generative models.
Using millions of text prompts from web-scale data and a filtering system, we generate a multi-view dataset comprising of 3 million multi-view videos. 

We utilize our synthetic multi-view dataset to train a 3D generative model capable of reconstructing 3D assets from single images in a feed-forward manner. Using the recently proposed Large Reconstruction Model (LRM)~\cite{hong2023lrm} as our starting point, we introduce a series of training strategies to stabilize the training process. These strategies assist the model in better learning from a substantial volume of synthetic multi-view training samples. Post-training, we further fine-tune with renderings from the 3D dataset, originally employed to fine-tune the video diffusion model, to further improve our 3D generative model. As a result, the proposed model, VFusion3D, can generate high-quality 3D assets from a single image with any viewing angles. We compare VFusion3D against several distillation-based and feed-forward 3D generative models using a user study and automated metrics. Finally, we undertake a comprehensive study to explore, analyze, and discuss topics on 3D data versus synthetic multi-view data, as well as the scaling trends of large 3D generative models.

\section{Related Work} 
\label{sec:related_work}

\paragraph{Text/Image-to-3D through Distillation or Reconstruction.} 
Learning to generate 3D assets from textual descriptions or single images presents a formidable challenge due to the limited availability of 3D-text paired data. DreamField~\cite{jain2022dreamfields} suggests using CLIP~\cite{radford2021learning} as a starting point, but textual supervision alone is not enough for 3D generation. DreamFusion~\cite{poole2022dreamfusion} optimizes an implicit 3D representation by distilling knowledge from 2D diffusion models~\cite{dhariwal2021diffusion,rombach2022high,saharia2022photorealistic} using a score-based distillation approach~\cite{poole2022dreamfusion}. Such distillation-based approaches has shown promising results and inspired subsequent research~\cite{lin2022magic3d,wang2023prolificdreamer,tang2023dreamgaussian,yi2023gaussiandreamer,lorraine2023att3d}. Furthermore, the combination of 2D diffusion models with 3D data to achieve both 3D consistency and visual pleasing is increasingly gaining popularity~\cite{liu2023zero,liu2023one,qian2023magic123,liu2023syncdreamer,li2023sweetdreamer,shi2023mvdream}. Concurrent research~\cite{melas2024im3d} suggests the reconstruction of 3D Gaussians~\cite{kerbl233d-gaussian} using multi-view outputs derived from fine-tuned video diffusion models~\cite{blattmann2023stable,kwak2023vivid}. Such reconstruction-based methods~\cite{liu2023one,liu2023one++,wu2023reconfusion,chan2023genvs} can also leverage multi-view data produced by other methods~\cite{liu2023zero,shi2023mvdream}.
With respect to distillation, our work also involves the distillation of 3D knowledge from video diffusion models. However, we distill knowledge from video diffusion models in an explicit way, eliminating the need for score distillation sampling~\cite{poole2022dreamfusion}.

\paragraph{Feed-forward 3D Generative Models.}
In the fields of 3D reconstruction and 3D generation, a prominent trend involves learning directly from 3D datasets~\cite{yu2021pixelnerf,erkocc2023hyperdiffusion,szymanowicz2023viewset, szymanowicz2023splatter,ren2023xcube,hong2023lrm, lorraine2023att3d,xu2024agg}. Pioneering works~\cite{jun2023shap,erkocc2023hyperdiffusion,szymanowicz2023viewset} attempted to learn 3D representations directly from 3D~\cite{deitke2023objaverse} and multi-view~\cite{yu2023mvimgnet} data, but these approaches faced limitations in performance due to low-quality training data or small-scale training data. A series of recent studies have suggested using a tri-plane~\cite{chan2022efficient} as the 3D representation, demonstrating that it can scale to larger datasets. These methods train large-scale models to generate a 3D representation from various inputs, including single images~\cite{hong2023lrm}, textual information~\cite{li2023instant3d}, posed multi-view images~\cite{xu2023dmv3d}, and unposed multi-view images~\cite{wang2023pf}. Typically, these large models~\cite{mercier2024hexagen3d} learn to predict a tri-plane representation, which is subsequently converted into a NeRF or combined with optimized point generator for Gaussian Splatting~\cite{zou2023triplane}. It is also possible to generate 3D Gaussians from multi-view inputs. Then, during inference time, it can employ off-the-shelf text-to-3D~\cite{wang2023imagedream} or image-to-3D models~\cite{shi2023mvdream} to generate multi-view inputs, as demonstrated in concurrent work~\cite{tang2024lgm}. In this study, we adopt the most general framework,\ie, LRM~\cite{openlrm,hong2023lrm}, without architectural modifications. Our objective is not to develop a new architecture. Instead, we strive to enhance the suitability and scalability of general models for training on large-scale synthetic multi-view images.

\section{Method}
Section~\ref{subsection:preliminaries} provides the preliminaries on the EMU Video~\cite{girdhar2023emuvideo} and LRM~\cite{hong2023lrm}. Our approach for transferring a video diffusion model into a 3D multi-view data engine is presented in Section~\ref{subsection:emuvideo}. The process of gradually improving the LRM into our VFusion3D is described in Section~\ref{subsection:f3d100k}. Figure~\ref{fig:overall} illustrates our pipeline.

\subsection{Preliminaries}
\label{subsection:preliminaries}

\paragraph{EMU Video.} EMU Video~\cite{girdhar2023emuvideo} is a video diffusion model that builds upon a pre-trained text-to-image diffusion model, EMU~\cite{dai2023emu}. It initializes its weights from the EMU model and adds new learnable temporal parameters. EMU Video is conditioned on both a text prompt and an image prompt, where the image prompt serves as the first frame and can either be provided or generated.
The EMU model is trained on 1.1 billion image-text pairs, and EMU Video is further trained on an additional 34 million video-text pairs. EMU Video is capable of generating high-quality and temporally consistent videos with up to 16 frames.

\paragraph{Large Reconstruction Model.} LRM is a large feed-forward model for single image 3D reconstruction.  It initially utilizes a pre-trained vision transformer, DINO~\cite{caron2021emerging}, to encode image features. These features are subsequently used by an image-to-triplane decoder module via cross-attention mechanisms. Camera parameters are also sent to the image-to-triplane decoder after undergoing processing through a small camera embedding network. The tri-plane representation, predicted by the image-to-triplane decoder, is upsampled and reshaped to query 3D point features. These features are then input into a multi-layer perception module to predict RGB and density for volumetric rendering. The training is conducted on 3D multi-view datasets~\cite{deitke2023objaverse,yu2023mvimgnet} and is supervised using multi-view images with LPIPS loss~\cite{zhang2018perceptual} and MSE loss.

\subsection{Video Diffusion Models as 3D Multi-view  Data Generator}
\label{subsection:emuvideo}

\paragraph{Prompt Collection and Filtering.} We gather object-centric prompts from web-scale datasets and 3D captions. Specifically, we collect textual captions from a large text-image dataset and employ a Llama2-13B~\cite{touvron2023llama} with a specially designed prompting template. This template incorporates in-context examples to retain those captions that describe a single object. Additionally, we run Cap3D~\cite{luo2023scalable} with a Llama2-70B on multi-view images rendered from some 3D data to collect captions of 3D objects. In total, we collect 4M prompts.

\paragraph{EMU Video Fine-tuning.}
Our aim is to enable the pre-trained EMU Video model to generate multi-view videos, which will showcase a camera rotating around a 3D object.
To accomplish this, we utilize an internal dataset of 100K 3D objects crafted by 3D modeling artists, akin to the 3D data collected in Objaverse~\cite{deitke2023objaverse}.
For each asset, we render 16 views with a random elevation within the range of $[0,\pi/4]$. The camera is positioned at uniform intervals ($2\pi / 16$ radians) around the object with a constant distance.

These rendered videos are used to fine-tune the EMU Video model. As all videos follow a similar camera trajectory, we do not send the camera parameters into the video diffusion model. Instead, we maintain a fixed camera distance and orientation, randomizing only the elevation. Despite the absence of explicit camera parameters, the first frame (the rendered image with 0 azimuth) can still exhibit varying viewing directions, enabling the EMU Video model to handle condition images with different viewing angles.
The EMU Video model is fine-tuned to generate a sequence of views that follows this distribution. The fine-tuning of the EMU Video model is conducted using a standard diffusion model training loss. We freeze all parameters, except for the temporal convolutional and attention layers to ensure that the fine-tuning does not degrade the visual quality of generation. 

\begin{figure}[t]
     \centering
     \includegraphics[width = \textwidth]
     {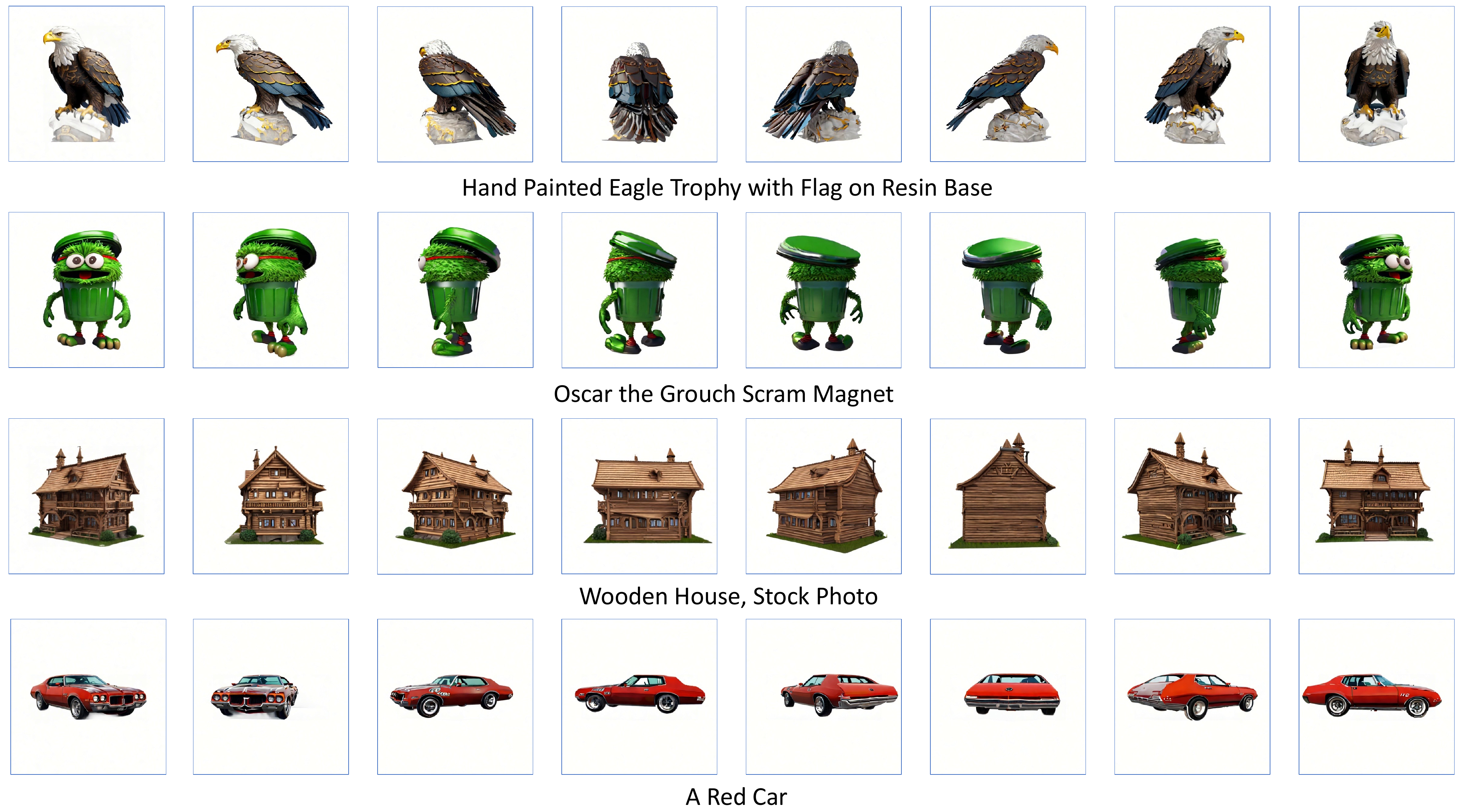}
    \caption{\textbf{Samples of multi-view sequences generated by fine-tuned video diffusion model}. After fine-tuning with 100K 3D data, the video diffusion model can produce high-quality multi-view sequences, thus functioning as a multi-view data generation engine. The inputs are the initial image and the textual prompt. The last row shows a failure case, for which we have designed a filter to remove such data.
     }
     \label{fig:emu_video_results}
\end{figure}

\paragraph{Post Processing and Metadata Preparation.}
We generated a total of 4 million videos, the majority of which demonstrate 3D-consistency and visual appeal. However, a subset exhibits lower quality and less 3D-consistency, prompting us to design a filter to selectively retain only those of superior quality. To achieve this, we manually labeled 2000 videos, with approximately 1200 classified as high quality. We utilized DINO~\cite{caron2021emerging} to extract features from 8 frames, uniformly distributed throughout the video. We subsequently trained a Support Vector Machine to classify video quality based on the averaged DINO features. As a result, we retained 2.7M videos for our synthetic dataset. Examples of generated multi-view sequences and data filtering are shown in Figure~\ref{fig:emu_video_results}.

The generated multi-view videos follow a similar camera trajectory to cover a 360$^\circ$ azimuth range, but the elevation depends on the generated condition image, which is less controllable. Therefore, it is necessary to label the elevation for multi-view videos. We used 100K 3D data with ground truth elevation to train our elevation estimator. Similar to the filtering process, we used DINO to extract features from 4 uniform views and averaged their features. A 2-layer MLP was trained on top of this as an elevation predictor. We then use it to label the elevation of all generated multi-view videos.

\subsection{VFusion3D}
\label{subsection:f3d100k}
VFusion3D adheres to the architecture of LRM, which is considered the most general method yet for feed-forward 3D creation. We pinpoint the ideal training strategies and formulate a recipe that enhances suitability and scalability when learning from synthetic multi-view data. Our ultimate objective is to establish learning from video diffusion generated multi-view data as a potential paradigm for training foundation 3D models.

\paragraph{Improved Training Strategies for Synthetic Data.}
The original training setting of LRM was designed to work with Objaverse~\cite{deitke2023objaverse} and MVImageNet~\cite{yu2023mvimgnet} data. However, we found that this setup was not entirely suitable for direct application to our synthetic data, as synthetic data tends to be noisier and does not always maintain full 3D consistency. 
To stabilize and improve the learning process on synthetic multi-view data, we implemented a series of strategies.

\noindent $\bullet$ \textit{\textbf{A Multi-stage Training Recipe.}}
LRM employs a patch rendering strategy, which renders a small $128 \times 128$ resolution patch from a random rendering resolution that ranges from 128 to 384. However, this setting can potentially cause instability in the learning process, particularly in the early stages, and may lead to incorrect optimization directions, resulting in predictions with checkerboard effects.
To address this, we suggest to use a multi-stage learning process. This involves gradually increasing the rendering resolution, thereby preparing the model for higher rendering resolutions at each stage. In practice, we use rendering resolutions of 128, 192, 256, 320, and 384. Each stage is trained for 5 epochs, with the exception of the final stage, which is trained for 10 epochs. After training on the 192 resolution stage, we reset the learning rate to the initial learning rate. This adjustment allows the model to better learn from later stages.

\noindent $\bullet$ \textbf{\textit{Image-level Supervision Instead of Pixel-level Supervision.} }Pixel-level losses, such as L1 and L2, are commonly used. However, they may not be suitable for synthetic multi-view data due to they are strict in pixel-level correspondence. Minor inconsistencies in synthetic multi-view images are inevitable, and when trained with pixel-level loss, these can lead to improper optimization and blurry results.
On the other hand, image-level loss methods, such as LPIPS~\cite{zhang2018perceptual}, are less stringent as they operate in feature spaces. This inherent flexibility has the potential to strengthen our model, as it accommodates minor inconsistencies in synthetic multi-view images without compromising the optimization process.

 \begin{figure}[!h]
     \centering
     \includegraphics[width = \textwidth]
     {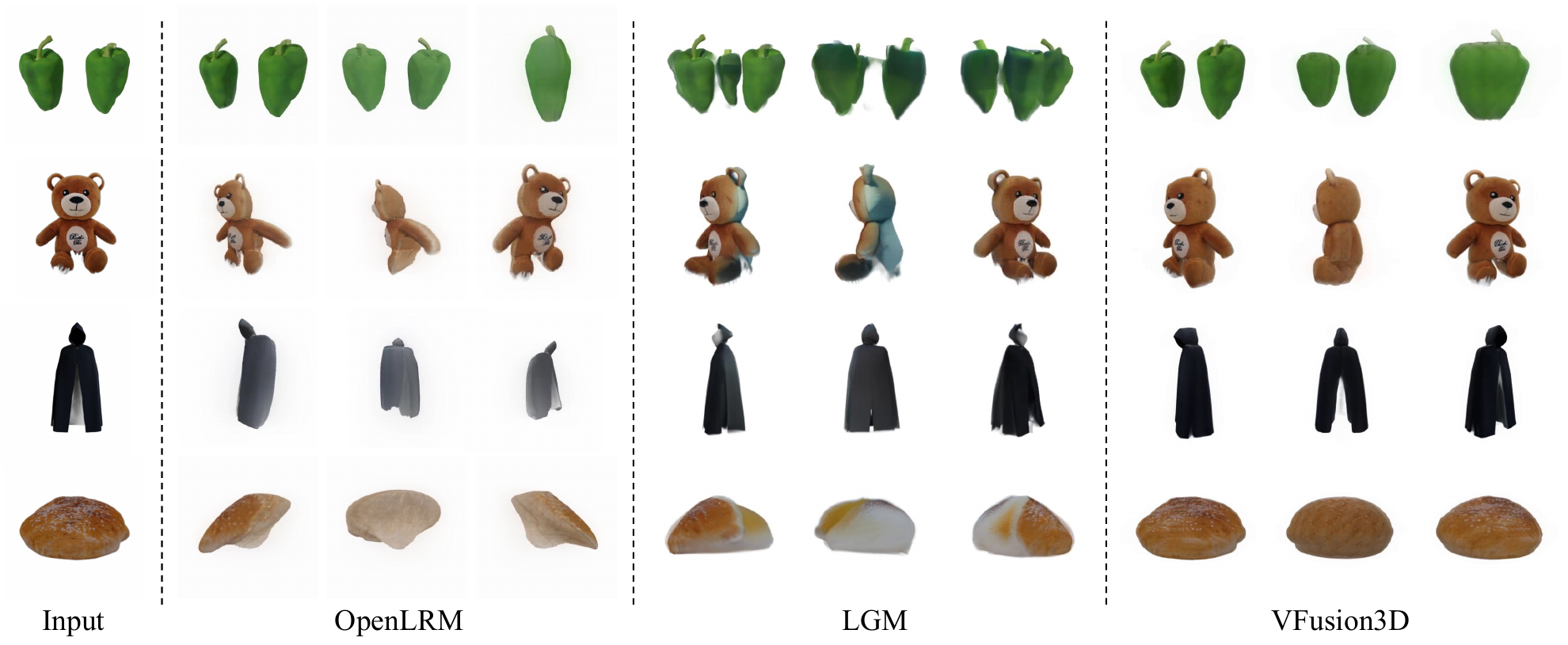}
\caption{\textbf{Qualitative results on single image 3D reconstruction}. VFusion3D successfully reconstructs 3D objects with strong 3D consistency in both shape and color. In contrast, OpenLRM sometimes fails to infer a reasonable shape, and LGM alters the color of unseen parts.}
     \label{fig:result_rec}
\end{figure}

\noindent $\bullet$ \textit{\textbf{Opacity Loss.}}
In synthetic multi-view data, small colored patches occasionally appear in images that should have a white background. To address this, we run a saliency detection model~\cite{Qin_2020_PR} to obtain the masks of central objects in our synthetic data. This allows us to apply an opacity loss, which helps maintain focus on the foreground object. Additionally, extra opacity supervision can facilitate the training process, resulting in stronger models.
We employ a general method to extract these masks and introduce an opacity loss between rendering density and masks. Combined with our strategy on no pixel-level losses, our approaches efficiently mitigate noises from small color patches in the background.

\begin{table}[!htb]
  \centering
  \fontsize{10}{3}\selectfont
  \setlength{\tabcolsep}{12pt} 
  \begin{tabular}{c|cc}
    \toprule
      Method & CLIP Text Similarity $\uparrow$ & CLIP Image Similarity $\uparrow$ \\
    \midrule
      OpenLRM & 0.234 & 0.793 \\
      LGM & 0.241 & 0.796 \\
      VFusion3D & \textbf{0.253} & \textbf{0.851} \\
    \bottomrule
  \end{tabular}
  \caption{\textbf{Comparative performance in image-to-3D tasks.} Overall, VFusion3D exhibits better performance in both text alignment and visual alignment.}
  \label{tab:imageto3d}
\end{table}

\noindent $\bullet$ \textit{\textbf{Camera Noise Injection.}}
For object observation, the camera trajectory of our synthetic multi-view data follows a fixed sequence of azimuth changes.
This could potentially restrict the model's generalization capability due to the limited viewing positions. To counteract this, we introduce camera noises during both the 3D data rendering and VFusion3D training stages. We apply a random minor offset (ranging from -0.05 to 0.05) during the rendering of multi-view images, which are subsequently used to fine-tune the EMU Video. Additionally, we infuse camera noise (varying from -0.02 to 0.02) into both the intrinsic and extrinsic matrices during the training process. This potentially enhances the model's robustness against incorrect camera parameters and improves its ability to generalize to different viewing angles.

\paragraph{Fine-tuning with 3D Data.}
Inspired by EMU~\cite{dai2023emu}, which employs a small number of visually striking images to fine-tune a pre-trained text-to-image diffusion model with a limited number of batches and iterations. We consider 3D data as visually striking and hypothesize that our model, which is pre-trained with synthetic multi-view sequences, could also benefit from exposure to some 3D samples. Following this procedure, we re-use the 100K data that was used in fine-tuning the EMU Video to fine-tune the VFusion3D model, which was pre-trained on synthetic data. We demonstrate that a combination of synthetic multi-view data and  3D data leads to the best model.

 \begin{figure}[!htb]
     \centering
     \includegraphics[width = \textwidth]
     {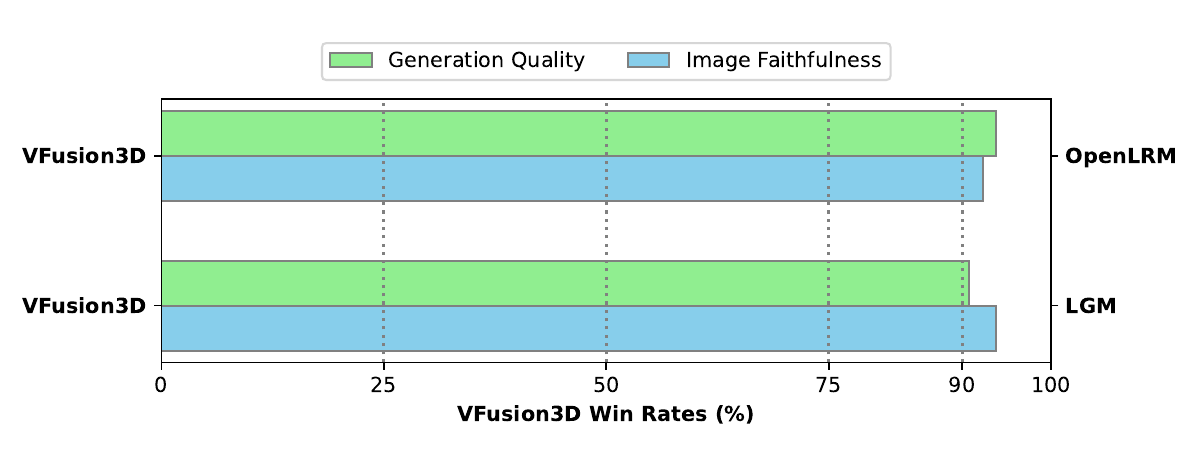}
\caption{\textbf{User study results comparing VFusion3D and previous work}. VFusion3D consistently outperforms previous works by considerable margins in both generation quality and image faithfulness.}
     \label{fig:winrate}
\end{figure}

 \section{Experiments}
\subsection{Implementation Details}
Our VFusion3D model is trained on 128 NVIDIA A100 (80G) GPUs, and the training process takes approximately 6 days to complete. We use a total batch size of 1024, with 4 multi-view images at a resolution of $128 \times 128$ used for supervision per batch, resulting in a total of 4096 multi-view images. The input image has a resolution of 512 $\times$ 512. The model is trained for 30 epochs with an initial learning rate of 4 $\times$ $10^{-4}$, following a cosine annealing schedule with a restart after first 10 epochs. The training begins with a warm-up of 3000 iterations, and the optimizer used is AdamW~\cite{loshchilov2017decoupled}. More details are presented in the Appendix~\ref{appendix}.

\subsection{Results and Comparisons}
\paragraph{Single Image 3D Reconstruction.}
\label{sec:rec}
We benchmark our model against recent large feed-forward methods, including OpenLRM-large~\cite{openlrm,hong2023lrm} and LGM~\cite{tang2024lgm}. For evaluation, we collect 25 diverse images that vary in style and shape. We employ BLIP-2~\cite{li2023blip} to obtain text captions. All results are rendered at a resolution of 384 $\times$ 384. We report both the CLIP text similarity scores and CLIP image similarity scores~\cite{radford2021learning}. Quantitative results are displayed in Table~\ref{tab:imageto3d} and qualitative results are presented in Figure~\ref{fig:result_rec}. Overall, VFusion3D exhibits superior performance in both text and image alignment, and it typically generates more visually appealing results with high image faithfulness.

\begin{figure}[!h]
     \centering
     \includegraphics[width = \textwidth]
     {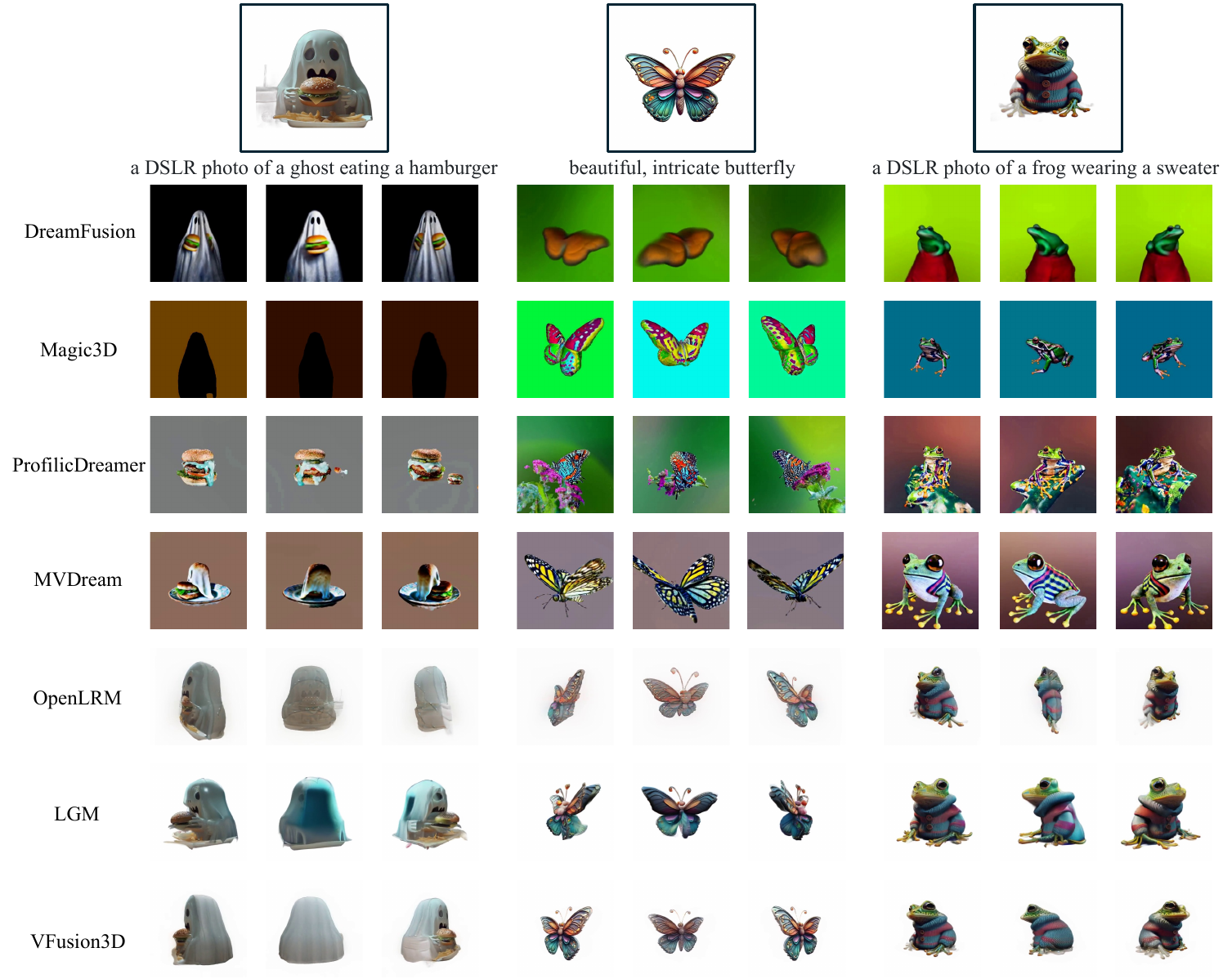}
    \caption{\textbf{Visual results of text-to-3D generation}. Among various methods, VFusion3D demonstrates strong performance. When compared with feed-forward reconstruction models, VFusion3D shows better image fidelity and 3D consistency.
     }
     \label{fig:result_text23d}
\end{figure}
  \begin{table}[!htb]
\centering
  \fontsize{10}{3}\selectfont
  \setlength{\tabcolsep}{8pt} 
  \begin{tabular}{c|cc}
    \toprule
      Method & CLIP Text Similarity $\uparrow$ & CLIP Image Similarity $\uparrow$ \\
\midrule 
DreamFusion & 0.261 & 0.640   \\
Magic3D     & 0.293 & 0.687  \\
ProlificDreamer     & \textbf{0.293} & 0.699   \\
MVDream     & 0.284 & 0.688  \\
\midrule 
OpenLRM & 0.255 & 0.826   \\
LGM & 0.270 & 0.832   \\
VFusion3D     & 0.272 & \textbf{0.899} \\
\bottomrule
\end{tabular}
\caption{\textbf{Comparisons of text-to-3D generation methods}. Text-to-3D methods typically exhibit stronger text alignment, whereas image-to-3D models often demonstrate better image alignment. VFusion3D yields the highest image similarity scores. }
\label{tab:textto3d}
\end{table}

\paragraph{Text-to-3D Generation.}
\label{sec:text-to-3d}
For text-to-3D generation, we employ EMU to generate images from text prompts, supporting text-image-3D reconstruction for our method. We use a common test set for evaluation, which contains 40 prompts from MVDream~\cite{shi2023mvdream}. We compare VFusion3D with DreamFusion~\cite{poole2022dreamfusion}, Magic3D~\cite{lin2022magic3d}, ProlificDreamer~\cite{wang2023prolificdreamer}, MVDream~\cite{shi2023mvdream}, OpenLRM~\cite{openlrm,hong2023lrm}, and LGM~\cite{tang2024lgm}. For text-to-3D models, we use rendered videos from the MVDream website for evaluation. Results are presented in Table~\ref{tab:textto3d}, and visual samples are shown in Figure~\ref{fig:result_text23d}. 
When considering feed-forward models, VFusion3D exhibits stronger image and text alignment. Qualitatively, VFusion3D shows superior results compared to other feed-forward models, especially in terms of both color and 3D shape consistency.

\begin{table}[t]
  \centering
  \fontsize{10}{3}\selectfont
  \setlength{\tabcolsep}{4pt} 
  \begin{tabular}{cc|cccc}
    \toprule
    \Rows{Num of Data} & \multicolumn{1}{c|}{Generated Videos} & \multicolumn{4}{c}{Trained Models} \\
    & Good Rate$\uparrow$ & SSIM$\uparrow$& LPIPS$\downarrow$ & Text Sim$\uparrow$ &Image Sim$\uparrow$  \\
    \midrule
    10K & 44.5$\%$   & 0.822 & 0.185 & 0.237 & 0.721\\
    50K & 52.1$\%$  & 0.823 & 0.184 & 0.247 & 0.753\\
    100K & \textbf{61.3$\%$}  & \textbf{0.824} & \textbf{0.182} &\textbf{0.250} & \textbf{0.759}\\
    \bottomrule
  \end{tabular}
  \caption{\textbf{Ablation study on the 3D data required for fine-tuning EMU video}. By utilizing more 3D data for fine-tuning EMU video, the ability to generate multi-view sequences in fine-tuned EMU video is improved.
  }
  \label{tab:ablation_finetunedata}
\end{table}

\paragraph{User Study.}
To assess the overall quality of the generated content and its faithfulness to the input images more accurately, we conducted a user study via Amazon Mechanical Turk. We presented users with two $360^{\circ}$ rendered videos - one produced by our VFusion3D model and another by a baseline model, and asked them to indicate a winner. A total of 65 videos (25 for 3D reconstruction, 40 for text-to-3D) were evaluated. Feedback was collected from 5 different users and we present the majority results in the form of a winner rate.
Figure~\ref{fig:winrate} displays our results. Our method surpasses SOTA baselines, demonstrating that VFusion3D aligns closely with the original image content and exhibits high visual quality.

\subsection{Ablation Study}
\label{sec:ablation}
We conduct an ablation study on several design choices, which include: (1) the specifications for fine-tuning the EMU Video model, (2) the training strategies proposed in VFusion3D, and (3) the number of 3D data for fine-tuning pre-trained VFusion3D. We use two evaluation sets for our study. First, we report SSIM and LPIPS for 3D data evaluation, along with CLIP text similarity score and CLIP image similarity score for text-to-3D evaluation. The 3D data evaluations were conducted on a test dataset comprising 500 unseen 3D models. From different viewing angles, we use one image as input, render 32 novel views, and compare them against their respective ground truths. The text-to-3D evaluation follows the setting of text-to-3D experiments as in section~\ref{sec:text-to-3d}. The VFusion3D model used in the ablation study~\ref{sec:ablation} and analysis~\ref{sec:analysis} is a variant that processes input images with a resolution of 224 $\times$ 224.

\paragraph{How Much 3D Data are Needed for Fine-tuning Video Diffusion Models?}
We explore the amount of 3D data required for fine-tuning EMU Video. Specifically, we experiment with 10K, 50K, and 100K 3D data. Table~\ref{tab:ablation_finetunedata} presents the results of both the good classification rate of multi-view sequences generated by the fine-tuned EMU Video and the performance of VFusion3D models trained on the synthetic datasets from the respective fine-tuned EMU Video. We generate 500K multi-view sequences before filtering for training VFusion3D variants. As expected, using more 3D data enhances the multi-view sequence generation ability of EMU Video without compromising visual quality. This suggests that our approach could also scale with the collection of more 3D data.

\paragraph{Effect of Improved Training Strategies.}
We evaluate the impact of the proposed training strategies incorporated in VFusion3D.  We use 500K synthetic data only for training ablation variants. The results of these variants are presented in Table~\ref{tab:ablation_component}.

Our findings include: (1) The multi-stage training recipe significantly improves the results by stabilizing the training process. (2) Without image-level supervision, such as MSE, we notice a small improvement across all metrics, as well as in the visual results. (3) The inclusion of opacity loss can further enhance performance. (4) While the injection of camera noises does not improve performance according to metrics, further testing reveals that this variant does enhance the model's robustness. It enables the model to handle a wider range of potentially inaccurate camera matrices. Therefore, the decision to apply it becomes a matter of trade-off.

\begin{table}[t]
  \centering
  \fontsize{10}{3}\selectfont
    \setlength{\tabcolsep}{3pt} 
    \begin{tabular}{lcccc}
    \toprule
     Components & SSIM$\uparrow$& LPIPS$\downarrow$ & CLIP Text Sim$\uparrow$ &  CLIP Image Sim$\uparrow$  \cr
    \midrule
    Baseline &  0.826& 0.206& 0.223 & 0.712\cr
    \midrule
    + multi-stage training& 0.829& 0.168 & 0.249 & 0.801  \cr
    + no pixel-level loss& \textbf{0.831}&\textbf{0.167} & \textbf{0.257} & 0.798  \cr
    + opacity supervision&\textbf{0.831}& \textbf{0.167}& 0.256 & \textbf{0.802}  \cr
    + camera noise& 0.830& 0.169 & 0.252 & 0.800 \cr
    \bottomrule
    \end{tabular}
    \caption{\textbf{Ablation study on improved training strategies}. We sequentially incorporate proposed strategies for synthetic data, and the effect of each is validated.}
     \label{tab:ablation_component}
\end{table}

\begin{table}[t]
  \centering
  \fontsize{10}{3}\selectfont
    \setlength{\tabcolsep}{6pt} 
    \begin{tabular}{ccccc}
    \toprule
     Num of data & SSIM$\uparrow$& LPIPS$\downarrow$ & CLIP Text Sim$\uparrow$  & CLIP Image Sim$\uparrow$ \cr
    \midrule
    No fine-tune & 0.832& 0.160 & 0.261 & 0.839 \cr
    \midrule
    10K& 0.835& 0.153 & 0.261 & 0.838  \cr
    30K& 0.837& 0.149 & 0.262 & \textbf{0.846}  \cr
    50K& 0.839& 0.147 & 0.262 & 0.834  \cr
    50K (random)& 0.840& 0.146 & 0.261 & 0.836 \cr
    50K (16-views)&  0.838& 0.147 & 0.262 & 0.835  \cr
    100K& \textbf{0.842}& \textbf{0.143}& \textbf{0.266} & 0.836 \cr
    \bottomrule
    \end{tabular}
    \caption{\textbf{Ablation study on settings of 3D data fine-tuning}. In general, the use of more 3D data leads to improved performance, irrespective of other settings.}
     \label{tab:ablation_emufinetune}
\end{table}

\paragraph{Settings of 3D Data Fine-tuning.}
We investigate the settings of 3D data fine-tuning, including sizes, data selections, and number of views. We evaluate datasets of varying sizes - 10K, 30K, 50K, and 100K - where the data is selected based on aesthetic score rankings, from high to low. To determine the aesthetic scores of 3D data, we train an aesthetic score predictor on top of a pre-trained 2D aesthetic score predictor. This predictor uses CLIP to extract features and an MLP for prediction. We average the CLIP features extracted from multi-view images to make it applicable to 3D data. To ascertain whether aesthetic score ranking is beneficial, we further draw 50K 3D data randomly from the 100k 3D data as a variant. By default, all 3D data are rendered to 32 views, but we also explore a variant that only uses 16 views.

Table~\ref{tab:ablation_emufinetune} presents the quantitative results. The conclusions drawn are as follows: (1) Fine-tuning with more 3D data results in stronger models. (2) Unlike 2D images, aesthetic scores are unnecessary in selecting 3D training data. (3) Even with only 16 views, VFusion3D, when trained with a large number of synthetic data, already learns how to interpolate effectively between different viewing angles. Fine-tuning with 32 views only leads to marginal improvements.

\subsection{Analysis and Discussion}
\label{sec:analysis}
In this analysis, our objective is to delve into the intricacies of synthetic multi-view data, exploring their benefits, limitations, and potential in comparison to 3D data. Given our capacity to generate a significant amount of synthetic multi-view data, which supports the data requirements for training foundation 3D generative models, we also present scaling trends of 3D generative models. The evaluation protocol in this section adheres to the ablation study~\ref{sec:ablation}.

\paragraph{3D Data \vs synthetic multi-view Data.}
We aim to investigate the performance of models trained on  3D data, synthetic data, and a combination of both (pre-training with synthetic, then fine-tuning with 3D data). For 3D data, we train a VFusion3D model using 150K 3D data from our internal dataset. This includes the same 100K 3D data that were used in the fine-tuning process of the EMU Video. Table~\ref{tab:analysis_data} presents the quantitative results. It shows that 3D data is more efficient than synthetic multi-view data in teaching the model to reconstruct common objects. Training with 100K 3D data points already matches the performance of 2.7M synthetic data points. However, learning with a limited number of data cannot generalize to uncommon objects or scenes, where large-scale synthetic data provides strong performance in generalization. An additional advantage of synthetic data is that it can be further combined with 3D data fine-tuning to achieve optimal performance.
\begin{table}[t]
  \centering
  \fontsize{10}{3}\selectfont
    \setlength{\tabcolsep}{4pt} 
    \begin{tabular}{ccc|cc}
    \toprule
     Data & SSIM$\uparrow$& LPIPS$\downarrow$ & CLIP Text Sim$\uparrow$ & CLIP Image Sim$\uparrow$ \cr
    \midrule
    3D data &  0.839& 0.161 & 0.205 & 0.631  \cr
    Synthetic data&  0.832& 0.160 & 0.261 & \textbf{0.839}  \cr
    Both 3D data &  \textbf{0.842}& \textbf{0.143 }& \textbf{0.266}  & 0.836 \cr
    \bottomrule
    \end{tabular}
     \caption{\textbf{Analysis on 3D data \vs Synthetic Multi-view Data}. 3D data is more efficient than synthetic data when it comes to learning to reconstruct common objects. However, large-scale synthetic multi-view data offers the ability to generalize to unusual objects and scenes. Combining the two can yield the best performance.}
     \label{tab:analysis_data}
\end{table}

\begin{figure}[!ht]
    \centering
    \begin{subfigure}[b]{0.48\textwidth}
        \includegraphics[width=\textwidth]{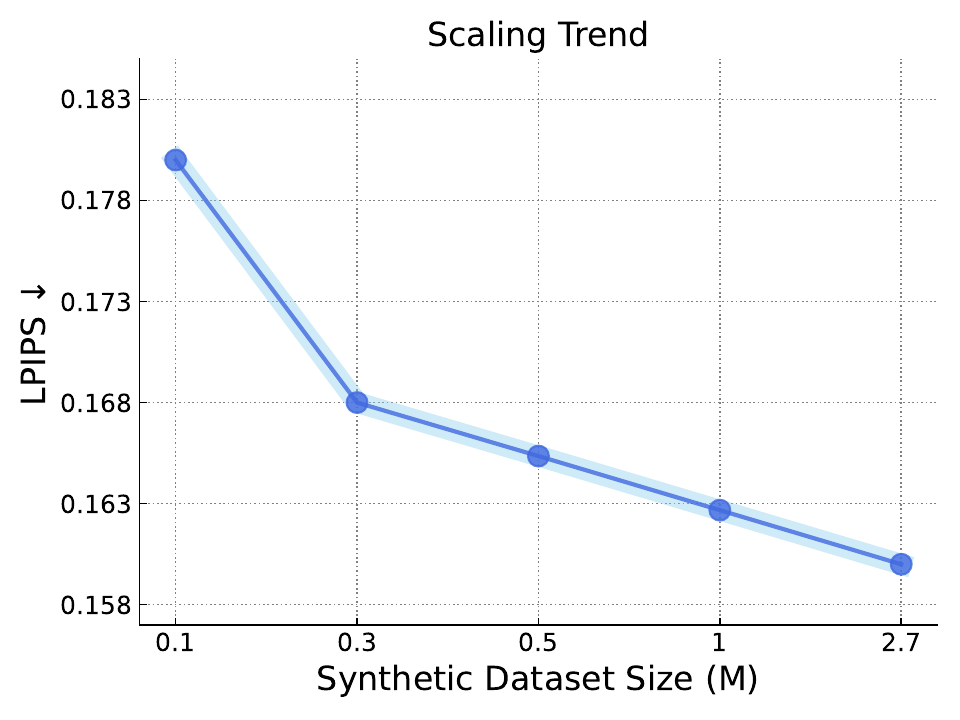}
    \end{subfigure}
    \hfill
    \begin{subfigure}[b]{0.48\textwidth}
        \includegraphics[width=\textwidth]{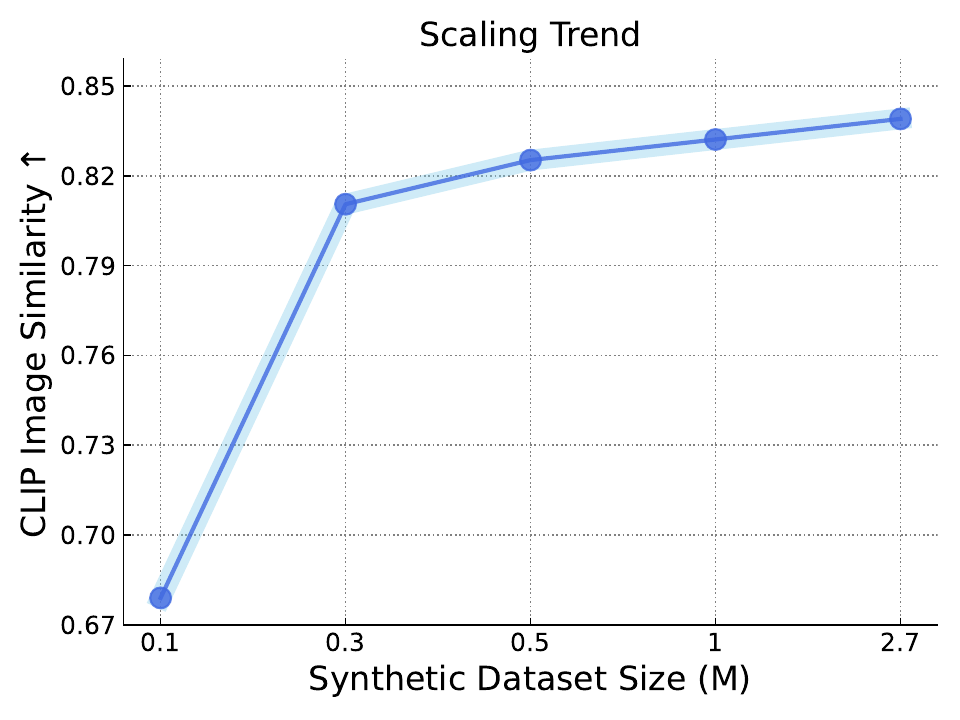}
    \end{subfigure}
\caption{\textbf{Scaling trends of VFusion3D on synthetic data}. The left and right figures display the LPIPS and CLIP image similarity scores in relation to the dataset size, respectively. The generation quality consistently improves as the dataset size increases.}
    \label{fig:scale}
\end{figure}  

\paragraph{Scaling Trends for 3D Generative Models.}
By maintaining a fixed model architecture, we examine the impact of varying training dataset sizes, ranging from 100K, 300K, 500K, 1M, to 2.7M. Our objective is to illustrate scaling trends that shed light on scalable 3D generative modeling. Figure~\ref{fig:scale} presents these trends. We observe that the generation quality, as measured by LPIPS and CLIP image similarity scores, consistently imrpoves with the size of the synthetic dataset. Given our ability to generate an unlimited amount of synthetic data, this makes our approach highly scalable.

Furthermore, our approach can also scale and improve with several other factors. These include the development of stronger video diffusion models, the availability of more 3D data for fine-tuning the video diffusion model and the pre-trained 3D generative model, and the advancement of large 3D feed-forward generative models. All these factors contribute to the scalability of our model, positioning it as a promising avenue for foundation 3D generative models.

\paragraph{Limitations.}
The fine-tuned video diffusion model is less effective at generating multi-view sequences of specific objects, such as vehicles like cars, bicycles, and motorcycles, and text-related content. Our filtering system excludes most of these less-than-ideal results; however, this could potentially affect the performance of the trained VFusion3D model due to an insufficient amount of data related to vehicles and texts. This is a shortcoming inherited from the pre-trained video diffusion model. With the development of progressively stronger video diffusion models, this limitation should be mitigated.

\section{Conclusion}
This paper leverages a video diffusion model as a multi-view data generator, thereby facilitating the learning of scalable 3D generative models. We have showcased the potential of video diffusion models to function as a multi-view data engine, capable of generating an even infinite scale of synthetic data to support scalable training. Our proposed model, VFusion3D, which learns from synthetic data, has shown superior performance in the generation of 3D assets. Beyond its current state, VFusion3D is highly scalable and can scale with both the number of synthetic data and 3D data, paving new paths for 3D generative models. 

\section*{Acknowledgements}
Junlin Han is supported by Meta. We would like to thank Jianyuan Wang, Luke Melas-Kyriazi, Yawar Siddiqui, Quankai Gao, Yanir Kleiman, Roman Shapovalov, Natalia Neverova, and Andrea Vedaldi for the insightful discussions and invaluable support.

\bibliographystyle{splncs04}
\bibliography{main}

\begin{thebibliography}{10}
\providecommand{\url}[1]{\texttt{#1}}
\providecommand{\urlprefix}{URL }
\providecommand{\doi}[1]{https://doi.org/#1}

\bibitem{achiam2023gpt}
Achiam, J., Adler, S., Agarwal, S., Ahmad, L., Akkaya, I., Aleman, F.L., Almeida, D., Altenschmidt, J., Altman, S., Anadkat, S., et~al.: Gpt-4 technical report. arXiv preprint arXiv:2303.08774  (2023)

\bibitem{blattmann2023stable}
Blattmann, A., Dockhorn, T., Kulal, S., Mendelevitch, D., Kilian, M., Lorenz, D., Levi, Y., English, Z., Voleti, V., Letts, A., et~al.: Stable video diffusion: Scaling latent video diffusion models to large datasets. arXiv preprint arXiv:2311.15127  (2023)

\bibitem{videoworldsimulators2024}
Brooks, T., Peebles, B., Homes, C., DePue, W., Guo, Y., Jing, L., Schnurr, D., Taylor, J., Luhman, T., Luhman, E., Ng, C.W.Y., Wang, R., Ramesh, A.: Video generation models as world simulators  (2024), \url{https://openai.com/research/video-generation-models-as-world-simulators}

\bibitem{caron2021emerging}
Caron, M., Touvron, H., Misra, I., J\'egou, H., Mairal, J., Bojanowski, P., Joulin, A.: Emerging properties in self-supervised vision transformers. In: Proceedings of the International Conference on Computer Vision (ICCV) (2021)

\bibitem{chan2022efficient}
Chan, E.R., Lin, C.Z., Chan, M.A., Nagano, K., Pan, B., De~Mello, S., Gallo, O., Guibas, L.J., Tremblay, J., Khamis, S., et~al.: Efficient geometry-aware 3d generative adversarial networks. In: CVPR (2022)

\bibitem{chan2023genvs}
Chan, E.R., Nagano, K., Chan, M.A., Bergman, A.W., Park, J.J., Levy, A., Aittala, M., De~Mello, S., Karras, T., Wetzstein, G.: Genvs: Generative novel view synthesis with 3d-aware diffusion models (2023)

\bibitem{chang2015shapenet}
Chang, A.X., Funkhouser, T., Guibas, L., Hanrahan, P., Huang, Q., Li, Z., Savarese, S., Savva, M., Song, S., Su, H., et~al.: Shapenet: An information-rich 3d model repository. arXiv preprint arXiv:1512.03012  (2015)

\bibitem{dai2023emu}
Dai, X., Hou, J., Ma, C.Y., Tsai, S., Wang, J., Wang, R., Zhang, P., Vandenhende, S., Wang, X., Dubey, A., et~al.: Emu: Enhancing image generation models using photogenic needles in a haystack. arXiv preprint arXiv:2309.15807  (2023)

\bibitem{deitke2024objaversexl}
Deitke, M., Liu, R., Wallingford, M., Ngo, H., Michel, O., Kusupati, A., Fan, A., Laforte, C., Voleti, V., Gadre, S.Y., et~al.: Objaverse-xl: A universe of 10m+ 3d objects. Advances in Neural Information Processing Systems  \textbf{36} (2024)

\bibitem{deitke2023objaverse}
Deitke, M., Schwenk, D., Salvador, J., Weihs, L., Michel, O., VanderBilt, E., Schmidt, L., Ehsani, K., Kembhavi, A., Farhadi, A.: Objaverse: A universe of annotated 3d objects. In: Proceedings of the IEEE/CVF Conference on Computer Vision and Pattern Recognition. pp. 13142--13153 (2023)

\bibitem{dhariwal2021diffusion}
Dhariwal, P., Nichol, A.: Diffusion models beat gans on image synthesis. NeurIPS  (2021)

\bibitem{downs2022google}
Downs, L., Francis, A., Koenig, N., Kinman, B., Hickman, R., Reymann, K., McHugh, T.B., Vanhoucke, V.: Google scanned objects: A high-quality dataset of 3d scanned household items. In: 2022 International Conference on Robotics and Automation (ICRA). pp. 2553--2560. IEEE (2022)

\bibitem{erkocc2023hyperdiffusion}
Erko{\c{c}}, Z., Ma, F., Shan, Q., Nie{\ss}ner, M., Dai, A.: Hyperdiffusion: Generating implicit neural fields with weight-space diffusion. arXiv preprint arXiv:2303.17015  (2023)

\bibitem{girdhar2023emuvideo}
Girdhar, R., Singh, M., Brown, A., Duval, Q., Azadi, S., Rambhatla, S.S., Shah, A., Yin, X., Parikh, D., Misra, I.: Emu video: Factorizing text-to-video generation by explicit image conditioning. arXiv preprint arXiv:2311.10709  (2023)

\bibitem{hao2021gancraft}
Hao, Z., Mallya, A., Belongie, S., Liu, M.Y.: Gancraft: Unsupervised 3d neural rendering of minecraft worlds. In: Proceedings of the IEEE/CVF International Conference on Computer Vision. pp. 14072--14082 (2021)

\bibitem{openlrm}
He, Z., Wang, T.: Openlrm: Open-source large reconstruction models. \url{https://github.com/3DTopia/OpenLRM} (2023)

\bibitem{hong2023lrm}
Hong, Y., Zhang, K., Gu, J., Bi, S., Zhou, Y., Liu, D., Liu, F., Sunkavalli, K., Bui, T., Tan, H.: Lrm: Large reconstruction model for single image to 3d. ICLR  (2024)

\bibitem{jain2022dreamfields}
Jain, A., Mildenhall, B., Barron, J.T., Abbeel, P., Poole, B.: Zero-shot text-guided object generation with dream fields. In: CVPR (2022)

\bibitem{jun2023shap}
Jun, H., Nichol, A.: Shap-e: Generating conditional 3d implicit functions. arXiv preprint arXiv:2305.02463  (2023)

\bibitem{kant2024spad}
Kant, Y., Siarohin, A., Wu, Z., Vasilkovsky, M., Qian, G., Ren, J., Guler, R.A., Ghanem, B., Tulyakov, S., Gilitschenski, I.: Spad: Spatially aware multi-view diffusers. In: Proceedings of the IEEE/CVF Conference on Computer Vision and Pattern Recognition. pp. 10026--10038 (2024)

\bibitem{kerbl233d-gaussian}
Kerbl, B., Kopanas, G., Leimk{\"u}hler, T., Drettakis, G.: 3d gaussian splatting for real-time radiance field rendering. ACM Transactions on Graphics  \textbf{42}(4) (2023)

\bibitem{kolotouros2024dreamhuman}
Kolotouros, N., Alldieck, T., Zanfir, A., Bazavan, E., Fieraru, M., Sminchisescu, C.: Dreamhuman: Animatable 3d avatars from text. Advances in Neural Information Processing Systems  \textbf{36} (2024)

\bibitem{kwak2023vivid}
Kwak, J.g., Dong, E., Jin, Y., Ko, H., Mahajan, S., Yi, K.M.: Vivid-1-to-3: Novel view synthesis with video diffusion models. arXiv preprint arXiv:2312.01305  (2023)

\bibitem{li2022rt}
Li, C., Li, S., Zhao, Y., Zhu, W., Lin, Y.: Rt-nerf: Real-time on-device neural radiance fields towards immersive ar/vr rendering. In: Proceedings of the 41st IEEE/ACM International Conference on Computer-Aided Design. pp.~1--9 (2022)

\bibitem{li2023instant3d}
Li, J., Tan, H., Zhang, K., Xu, Z., Luan, F., Xu, Y., Hong, Y., Sunkavalli, K., Shakhnarovich, G., Bi, S.: Instant3d: Fast text-to-3d with sparse-view generation and large reconstruction model. ICLR  (2024)

\bibitem{li2023blip}
Li, J., Li, D., Savarese, S., Hoi, S.: Blip-2: Bootstrapping language-image pre-training with frozen image encoders and large language models. arXiv preprint arXiv:2301.12597  (2023)

\bibitem{li2023sweetdreamer}
Li, W., Chen, R., Chen, X., Tan, P.: Sweetdreamer: Aligning geometric priors in 2d diffusion for consistent text-to-3d. arXiv preprint arXiv:2310.02596  (2023)

\bibitem{lin2022magic3d}
Lin, C.H., Gao, J., Tang, L., Takikawa, T., Zeng, X., Huang, X., Kreis, K., Fidler, S., Liu, M.Y., Lin, T.Y.: Magic3{D}: High-resolution text-to-3d content creation. arXiv preprint arXiv:2211.10440  (2022)

\bibitem{liu2023one++}
Liu, M., Shi, R., Chen, L., Zhang, Z., Xu, C., Wei, X., Chen, H., Zeng, C., Gu, J., Su, H.: One-2-3-45++: Fast single image to 3d objects with consistent multi-view generation and 3d diffusion. arXiv preprint arXiv:2311.07885  (2023)

\bibitem{liu2023one}
Liu, M., Xu, C., Jin, H., Chen, L., Xu, Z., Su, H., et~al.: One-2-3-45: Any single image to 3d mesh in 45 seconds without per-shape optimization. arXiv preprint arXiv:2306.16928  (2023)

\bibitem{liu2023zero}
Liu, R., Wu, R., Van~Hoorick, B., Tokmakov, P., Zakharov, S., Vondrick, C.: Zero-1-to-3: Zero-shot one image to 3d object. In: Proceedings of the IEEE/CVF International Conference on Computer Vision. pp. 9298--9309 (2023)

\bibitem{liu2023syncdreamer}
Liu, Y., Lin, C., Zeng, Z., Long, X., Liu, L., Komura, T., Wang, W.: Syncdreamer: Generating multiview-consistent images from a single-view image. arXiv preprint arXiv:2309.03453  (2023)

\bibitem{lorensen1998marching}
Lorensen, W.E., Cline, H.E.: Marching cubes: A high resolution 3d surface construction algorithm. In: Seminal graphics: pioneering efforts that shaped the field, pp. 347--353 (1998)

\bibitem{lorraine2023att3d}
Lorraine, J., Xie, K., Zeng, X., Lin, C.H., Takikawa, T., Sharp, N., Lin, T.Y., Liu, M.Y., Fidler, S., Lucas, J.: Att3d: Amortized text-to-3d object synthesis. arXiv preprint arXiv:2306.07349  (2023)

\bibitem{loshchilov2017decoupled}
Loshchilov, I., Hutter, F.: Decoupled weight decay regularization. arXiv preprint arXiv:1711.05101  (2017)

\bibitem{luo2023scalable}
Luo, T., Rockwell, C., Lee, H., Johnson, J.: Scalable 3d captioning with pretrained models. arXiv preprint arXiv:2306.07279  (2023)

\bibitem{melas2024im3d}
Melas-Kyriazi, L., Laina, I., Rupprecht, C., Neverova, N., Vedaldi, A., Gafni, O., Kokkinos, F.: Im-3d: Iterative multiview diffusion and reconstruction for high-quality 3d generation. arXiv preprint arXiv:2402.08682  (2024)

\bibitem{mercier2024hexagen3d}
Mercier, A., Nakhli, R., Reddy, M., Yasarla, R., Cai, H., Porikli, F., Berger, G.: Hexagen3d: Stablediffusion is just one step away from fast and diverse text-to-3d generation. arXiv preprint arXiv:2401.07727  (2024)

\bibitem{mescheder2019occupancy}
Mescheder, L., Oechsle, M., Niemeyer, M., Nowozin, S., Geiger, A.: Occupancy networks: Learning 3d reconstruction in function space. In: Proceedings of the IEEE/CVF conference on computer vision and pattern recognition. pp. 4460--4470 (2019)

\bibitem{mildenhall2021nerf}
Mildenhall, B., Srinivasan, P.P., Tancik, M., Barron, J.T., Ramamoorthi, R., Ng, R.: Nerf: Representing scenes as neural radiance fields for view synthesis. Communications of the ACM  \textbf{65}(1),  99--106 (2021)

\bibitem{oquab2023dinov2}
Oquab, M., Darcet, T., Moutakanni, T., Vo, H., Szafraniec, M., Khalidov, V., Fernandez, P., Haziza, D., Massa, F., El-Nouby, A., et~al.: Dinov2: Learning robust visual features without supervision. arXiv preprint arXiv:2304.07193  (2023)

\bibitem{park2019deepsdf}
Park, J.J., Florence, P., Straub, J., Newcombe, R., Lovegrove, S.: Deepsdf: Learning continuous signed distance functions for shape representation. In: Proceedings of the IEEE/CVF conference on computer vision and pattern recognition. pp. 165--174 (2019)

\bibitem{peebles2023scalable}
Peebles, W., Xie, S.: Scalable diffusion models with transformers. In: Proceedings of the IEEE/CVF International Conference on Computer Vision. pp. 4195--4205 (2023)

\bibitem{poole2022dreamfusion}
Poole, B., Jain, A., Barron, J.T., Mildenhall, B.: Dream{F}usion: Text-to-3d using 2d diffusion. arXiv  (2022)

\bibitem{qian2023magic123}
Qian, G., Mai, J., Hamdi, A., Ren, J., Siarohin, A., Li, B., Lee, H.Y., Skorokhodov, I., Wonka, P., Tulyakov, S., et~al.: Magic123: One image to high-quality 3d object generation using both 2d and 3d diffusion priors. arXiv preprint arXiv:2306.17843  (2023)

\bibitem{qin2022}
Qin, X., Dai, H., Hu, X., Fan, D.P., Shao, L., Gool, L.V.: Highly accurate dichotomous image segmentation. In: ECCV (2022)

\bibitem{Qin_2020_PR}
Qin, X., Zhang, Z., Huang, C., Dehghan, M., Zaiane, O., Jagersand, M.: U2-net: Going deeper with nested u-structure for salient object detection. vol.~106, p. 107404 (2020)

\bibitem{radford2021learning}
Radford, A., Kim, J.W., Hallacy, C., Ramesh, A., Goh, G., Agarwal, S., Sastry, G., Askell, A., Mishkin, P., Clark, J., Krueger, G., Sutskever, I.: Learning transferable visual models from natural language supervision. In: ICML (2021)

\bibitem{ren2023xcube}
Ren, X., Huang, J., Zeng, X., Museth, K., Fidler, S., Williams, F.: Xcube (x3): Large-scale 3d generative modeling using sparse voxel hierarchies. arXiv preprint arXiv:2312.03806  (2023)

\bibitem{rombach2022high}
Rombach, R., Blattmann, A., Lorenz, D., Esser, P., Ommer, B.: High-resolution image synthesis with latent diffusion models. In: Proceedings of the IEEE/CVF conference on computer vision and pattern recognition. pp. 10684--10695 (2022)

\bibitem{saharia2022photorealistic}
Saharia, C., Chan, W., Saxena, S., Li, L., Whang, J., Denton, E., Ghasemipour, S.K.S., Ayan, B.K., Mahdavi, S.S., Lopes, R.G., Salimans, T., Ho, J., Fleet, D.J., Norouzi, M.: Photorealistic text-to-image diffusion models with deep language understanding. arXiv preprint arXiv:2205.11487  (2022)

\bibitem{shi2023mvdream}
Shi, Y., Wang, P., Ye, J., Long, M., Li, K., Yang, X.: Mvdream: Multi-view diffusion for 3d generation. arXiv preprint arXiv:2308.16512  (2023)

\bibitem{sun20233d}
Sun, C., Han, J., Deng, W., Wang, X., Qin, Z., Gould, S.: 3d-gpt: Procedural 3d modeling with large language models. arXiv preprint arXiv:2310.12945  (2023)

\bibitem{szymanowicz2023splatter}
Szymanowicz, S., Rupprecht, C., Vedaldi, A.: Splatter image: Ultra-fast single-view 3d reconstruction. arXiv preprint arXiv:2312.13150  (2023)

\bibitem{szymanowicz2023viewset}
Szymanowicz, S., Rupprecht, C., Vedaldi, A.: Viewset diffusion:(0-) image-conditioned 3d generative models from 2d data. arXiv preprint arXiv:2306.07881  (2023)

\bibitem{tang2024lgm}
Tang, J., Chen, Z., Chen, X., Wang, T., Zeng, G., Liu, Z.: Lgm: Large multi-view gaussian model for high-resolution 3d content creation. arXiv preprint arXiv:2402.05054  (2024)

\bibitem{tang2023dreamgaussian}
Tang, J., Ren, J., Zhou, H., Liu, Z., Zeng, G.: Dreamgaussian: Generative gaussian splatting for efficient 3d content creation. arXiv preprint arXiv:2309.16653  (2023)

\bibitem{tang2023MVDiffusion}
Tang, S., Zhang, F., Chen, J., Wang, P., Yasutaka, F.: Mvdiffusion: Enabling holistic multi-view image generation with correspondence-aware diffusion. arXiv preprint 2307.01097  (2023)

\bibitem{team2023gemini}
Team, G., Anil, R., Borgeaud, S., Wu, Y., Alayrac, J.B., Yu, J., Soricut, R., Schalkwyk, J., Dai, A.M., Hauth, A., et~al.: Gemini: a family of highly capable multimodal models. arXiv preprint arXiv:2312.11805  (2023)

\bibitem{touvron2023llama}
Touvron, H., Martin, L., Stone, K., Albert, P., Almahairi, A., Babaei, Y., Bashlykov, N., Batra, S., Bhargava, P., Bhosale, S., et~al.: Llama 2: Open foundation and fine-tuned chat models. arXiv preprint arXiv:2307.09288  (2023)

\bibitem{voleti2024sv3d}
Voleti, V., Yao, C.H., Boss, M., Letts, A., Pankratz, D., Tochilkin, D., Laforte, C., Rombach, R., Jampani, V.: Sv3d: Novel multi-view synthesis and 3d generation from a single image using latent video diffusion. arXiv preprint arXiv:2403.12008  (2024)

\bibitem{wang2023vggsfm}
Wang, J., Karaev, N., Rupprecht, C., Novotny, D.: Vggsfm: Visual geometry grounded deep structure from motion (2023)

\bibitem{wang2023imagedream}
Wang, P., Shi, Y.: Imagedream: Image-prompt multi-view diffusion for 3d generation. arXiv preprint arXiv:2312.02201  (2023)

\bibitem{wang2023pf}
Wang, P., Tan, H., Bi, S., Xu, Y., Luan, F., Sunkavalli, K., Wang, W., Xu, Z., Zhang, K.: Pf-lrm: Pose-free large reconstruction model for joint pose and shape prediction. ICLR  (2024)

\bibitem{wang2023prolificdreamer}
Wang, Z., Lu, C., Wang, Y., Bao, F., Li, C., Su, H., Zhu, J.: Prolificdreamer: High-fidelity and diverse text-to-3d generation with variational score distillation. arXiv preprint arXiv:2305.16213  (2023)

\bibitem{wu2023reconfusion}
Wu, R., Mildenhall, B., Henzler, P., Park, K., Gao, R., Watson, D., Srinivasan, P.P., Verbin, D., Barron, J.T., Poole, B., et~al.: Reconfusion: 3d reconstruction with diffusion priors. arXiv preprint arXiv:2312.02981  (2023)

\bibitem{xu2024agg}
Xu, D., Yuan, Y., Mardani, M., Liu, S., Song, J., Wang, Z., Vahdat, A.: Agg: Amortized generative 3d gaussians for single image to 3d. arXiv preprint arXiv:2401.04099  (2024)

\bibitem{xu2023dmv3d}
Xu, Y., Tan, H., Luan, F., Bi, S., Wang, P., Li, J., Shi, Z., Sunkavalli, K., Wetzstein, G., Xu, Z., et~al.: Dmv3d: Denoising multi-view diffusion using 3d large reconstruction model. ICLR  (2024)

\bibitem{yang2024consistnet}
Yang, J., Cheng, Z., Duan, Y., Ji, P., Li, H.: Consistnet: Enforcing 3d consistency for multi-view images diffusion. In: Proceedings of the IEEE/CVF Conference on Computer Vision and Pattern Recognition. pp. 7079--7088 (2024)

\bibitem{yi2023gaussiandreamer}
Yi, T., Fang, J., Wang, J., Wu, G., Xie, L., Zhang, X., Liu, W., Tian, Q., Wang, X.: Gaussiandreamer: Fast generation from text to 3d gaussians by bridging 2d and 3d diffusion models. arXiv preprint arXiv:2310.08529  (2023)

\bibitem{yu2021pixelnerf}
Yu, A., Ye, V., Tancik, M., Kanazawa, A.: pixelnerf: Neural radiance fields from one or few images. In: Proceedings of the IEEE/CVF Conference on Computer Vision and Pattern Recognition. pp. 4578--4587 (2021)

\bibitem{yu2023mvimgnet}
Yu, X., Xu, M., Zhang, Y., Liu, H., Ye, C., Wu, Y., Yan, Z., Zhu, C., Xiong, Z., Liang, T., et~al.: Mvimgnet: A large-scale dataset of multi-view images. In: Proceedings of the IEEE/CVF Conference on Computer Vision and Pattern Recognition. pp. 9150--9161 (2023)

\bibitem{zhang2018perceptual}
Zhang, R., Isola, P., Efros, A.A., Shechtman, E., Wang, O.: The unreasonable effectiveness of deep features as a perceptual metric. In: CVPR (2018)

\bibitem{zou2023triplane}
Zou, Z.X., Yu, Z., Guo, Y.C., Li, Y., Liang, D., Cao, Y.P., Zhang, S.H.: Triplane meets gaussian splatting: Fast and generalizable single-view 3d reconstruction with transformers. arXiv preprint arXiv:2312.09147  (2023)

\end{thebibliography}

\clearpage
\appendix
\section{arXiv Update Notes}
This arXiv v2 version is the ECCV 2024 camera-ready version. Significant updates have been made compared to the arXiv v1 version, including: (1) Updating all major results of VFusion3D, which now uses a resolution of 512x512 for input images during both training and inference, compared to the 224$\times$224 resolution used in v1. (2) Incorporating benchmarking results on GSO~\cite{downs2022google}. (3) Adding experiments on SfM for fine-tuned Emu Video generated multi-view sequences. (4) Adding discussions on potential solutions for our current limitations with updated references. 

\section{Training Details}
\label{appendix}
\paragraph{Emu Video Fine-tuning.}
Following the EMU Video~\cite{girdhar2023emuvideo}, we freeze the spatial convolutional and attention layers of Emu Video, while only fine-tuning the temporal layers. We use the standard diffusion loss for this fine-tuning process. The Emu Video model is fine-tuned over a period of 5 days using 80 A100 GPUs, with a total batch size of 240 and a learning rate of 1 $\times$ $10^{-5}$. Although the 3D consistency continues to improve with extended fine-tuning, we do not observe any decline in visual quality. One possible explanation is the static nature of the spatial layers and the image-conditioned network, which ensures that the generated 360$^\circ$ videos maintain high fidelity with the high-frequency texture components of the input.

\paragraph{VFusion3D.}
The architecture of VFusion3D is identical to that of LRM~\cite{hong2023lrm}, except for two minor modifications. The first is the use of DINOv2~\cite{oquab2023dinov2} instead of DINO~\cite{caron2021emerging} to enhance image feature extraction capabilities. The second modification is that we have reduced the number of NeRF MLP layers to 4 instead of 10, as empirically suggested by the LRM ablation study. 

In addition to the training details provided in the main paper, we use 0.95 as the second beta parameter of the AdamW optimizer~\cite{loshchilov2017decoupled}. We apply a gradient clipping of 1.0 and a weight decay of 0.05. This weight decay is only applied to weights that are neither bias nor part of normalization layers. We use Bfloat16 precision for training and inference.

\paragraph{Fine-tuning with 3D Data.}
We use 32 GPUs to fine-tune the pre-trained VFusion3D model with 3D data. At this stage, we also employ the L2 loss function for novel view supervision. The model undergoes fine-tuning with a dataset of 100K rendered multi-view images over 10 epochs, adhering to a cosine learning rate schedule. We set the initial learning rate as $1 \times 10^{-4}$. All other parameters remain consistent with the VFusion3D pre-training phase.

\section{Visualizations and Test-time Processing}

\paragraph{Video Comparison Results.}
We provide video comparison results in the project page that cover all the qualitative results presented in the main paper. Our project page also includes additional comparisons. These additional results are based on the Single Image 3D Reconstruction and Text-to-3D Generation experiments discussed in the main paper. All input images used were never seen by the model during training.

\paragraph{Test-time Processing.}
Following standard procedures~\cite{hong2023lrm,tang2024lgm}, we utilize a heuristic function to process all input images during testing. The initial steps involve eliminating the image background using rembg with isnet-general-use~\cite{qin2022}, then extracting the salient object. Following this, we adjust the size of the salient object to an appropriate scale and position it in the center of the input image.

\section{Benchmarking on Public Datasets}
We present a comparison using the GSO~\cite{downs2022google} dataset, from which we render 996 objects (excluding some redundant or very similar objects) with each object having 16 rendered views. These views are rendered with a fixed elevation angle of 20$^{\circ}$. The azimuth angle is sampled uniformly between 0$^{\circ}$ and 360$^{\circ}$. The input image is the first rendered view, namely, the view with an azimuth of 0$^{\circ}$.

Results are shown in Table~\ref{tab:gso}, which demonstrates that VFusion3D significantly outperforms the baselines by large margins. This further demonstrates that utilizing synthetic data to scale 3D generative models holds great promise.

\begin{table}[h]
  \centering
  \fontsize{10}{3}\selectfont
    \setlength{\tabcolsep}{6pt} 
    \begin{tabular}{cccc}
    \toprule
     Method & PSNR$\uparrow$  & SSIM$\uparrow$  & LPIPS$\downarrow$ \cr
    \midrule
    OpenLRM& 14.21& 0.820  & 0.226   \cr
    LGM& 13.53& 0.811 & 0.254  \cr
    VFusion3D & \textbf{20.89}& \textbf{0.848}& \textbf{0.127} \cr
    \bottomrule
    \end{tabular}
    \caption{Performance Comparison on the GSO Dataset: VFusion3D significantly outperforms the baselines by substantial margins.}
        \label{tab:gso}
\end{table}
 
\section{Discussions on Limitations and Potential Solutions}
\begin{figure}[t]
     \centering
     \includegraphics[width = \textwidth]
     {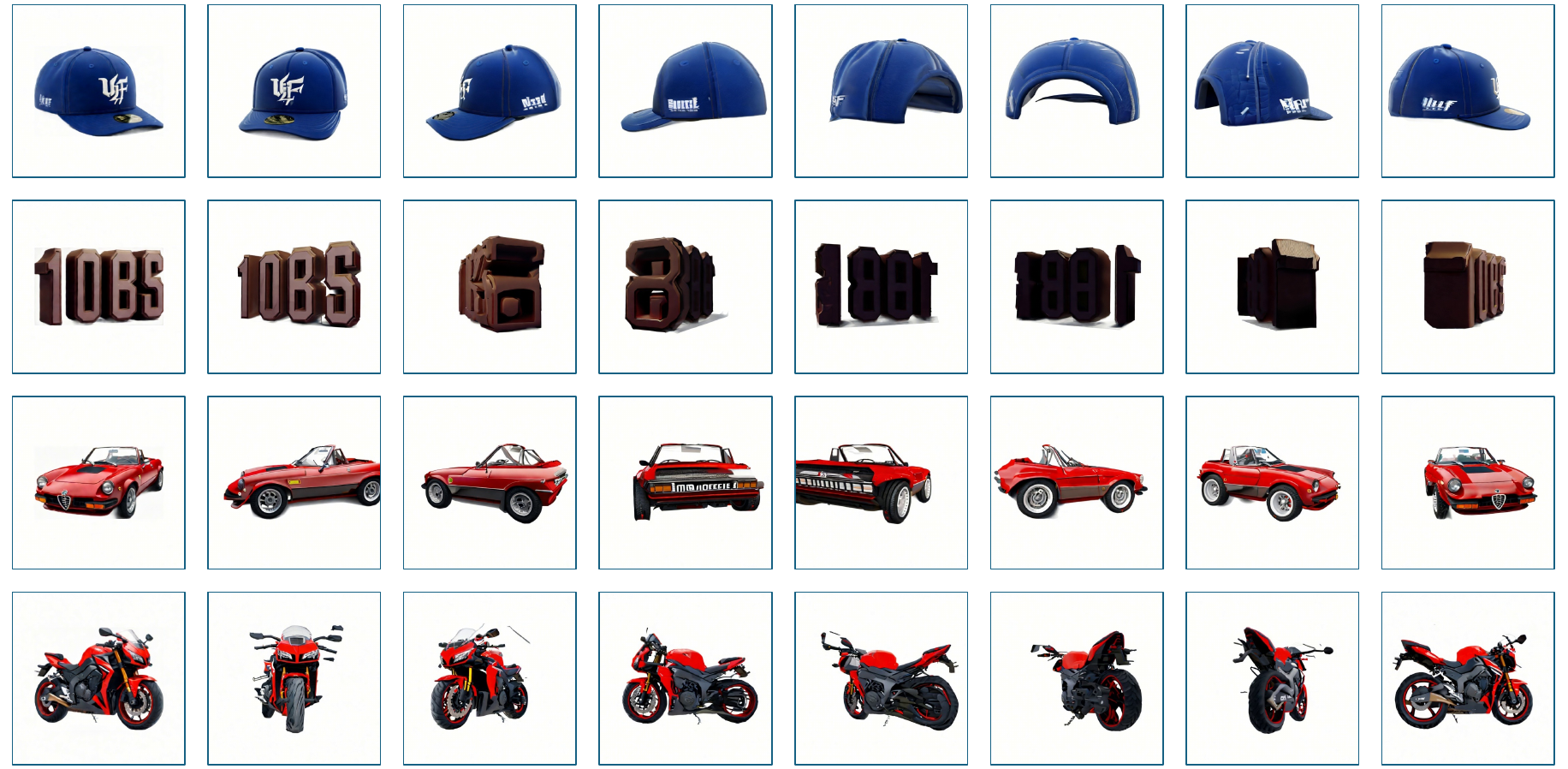}     
\caption{\textbf{Samples of failure cases generated by the fine-tuned video diffusion model}. In these instances, our fine-tuned video diffusion model struggles to generate high-quality multi-view sequences of text-related content and vehicles, resulting in distortions and 3D inconsistencies. Most of these failure cases are subsequently filtered out by the designed filter.}
     \label{fig:failure}
\end{figure}

The limitations section of the main paper presents that the fine-tuned video generator does not always yield flawless results. This is particularly noticeable in scenarios involving vehicles and texts, where the model sometimes generates multi-view results that lack 3D consistency. Additional examples of this are presented in Figure~\ref{fig:failure}.  

\begin{figure}[t]
     \centering
     \includegraphics[width = \textwidth]
     {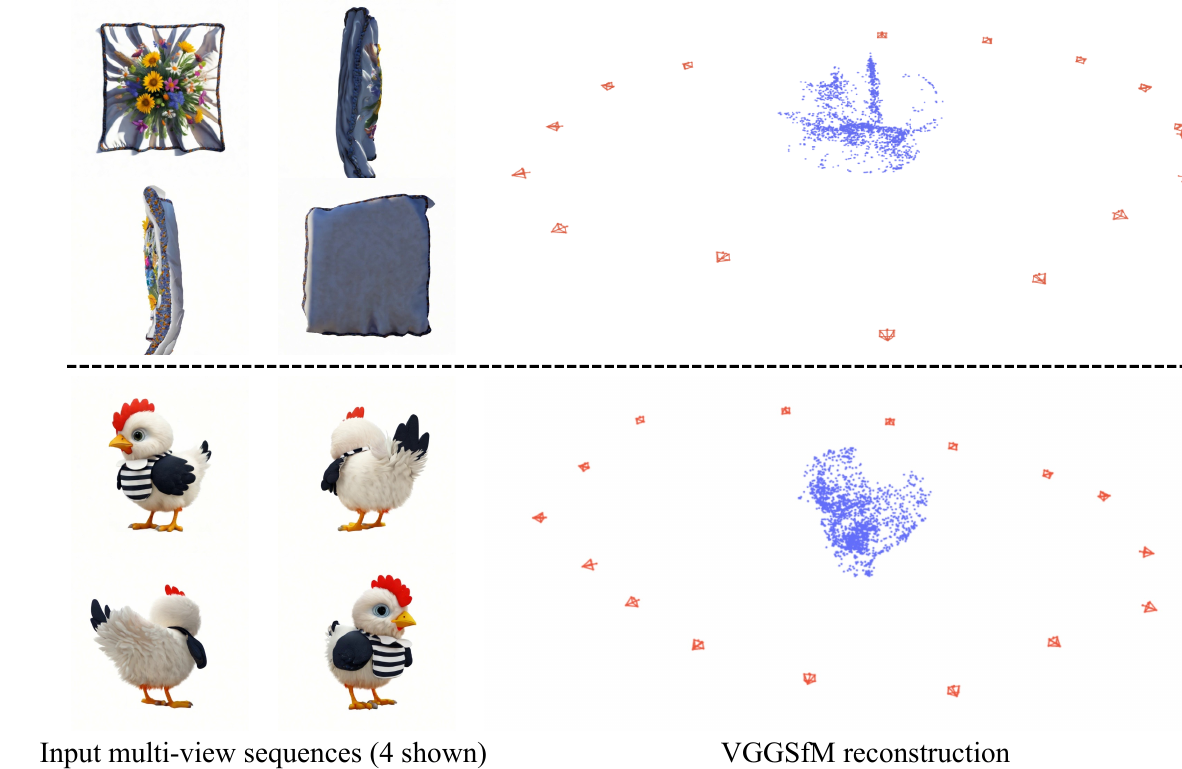}
    \caption{\textbf{VGGSfM reconstruction on all 16 generated frames}. Our fine-tuned EMU Video typically produces multi-view sequences with reasonable pixel-level correspondences.}
     \label{fig:vggsfm}
\end{figure}

One way to assess the failure rate and check the pixel-level multi-view correspondence is to run Structure from Motion (SfM) algorithms, which require pixel-level correspondences, on generated multi-view frames. We run VGGSfM~\cite{wang2023vggsfm} on 1000 fine-tuned EMU Video-generated multi-view sequences. Among these sequences, VGGSfM is able to obtain reasonable pose estimations (at least accurate poses for 8 views) over 70\% of the time. Examples are presented in Figure~\ref{fig:vggsfm}. This demonstrates that the generated sequences generally have reasonable pixel-level 3D correspondence, but there is significant room for further improvements. Enforcing stronger multi-view consistency constraints~\cite{voleti2024sv3d,kant2024spad,yang2024consistnet,tang2023MVDiffusion} during the video fine-tuning stage is a plausible solution. This could involve the use of epipolar constraints and 3D-aware Plücker ray embeddings.

\section{Conversation to Meshes}
\begin{figure}[t]
     \centering
     \includegraphics[width = \textwidth]
     {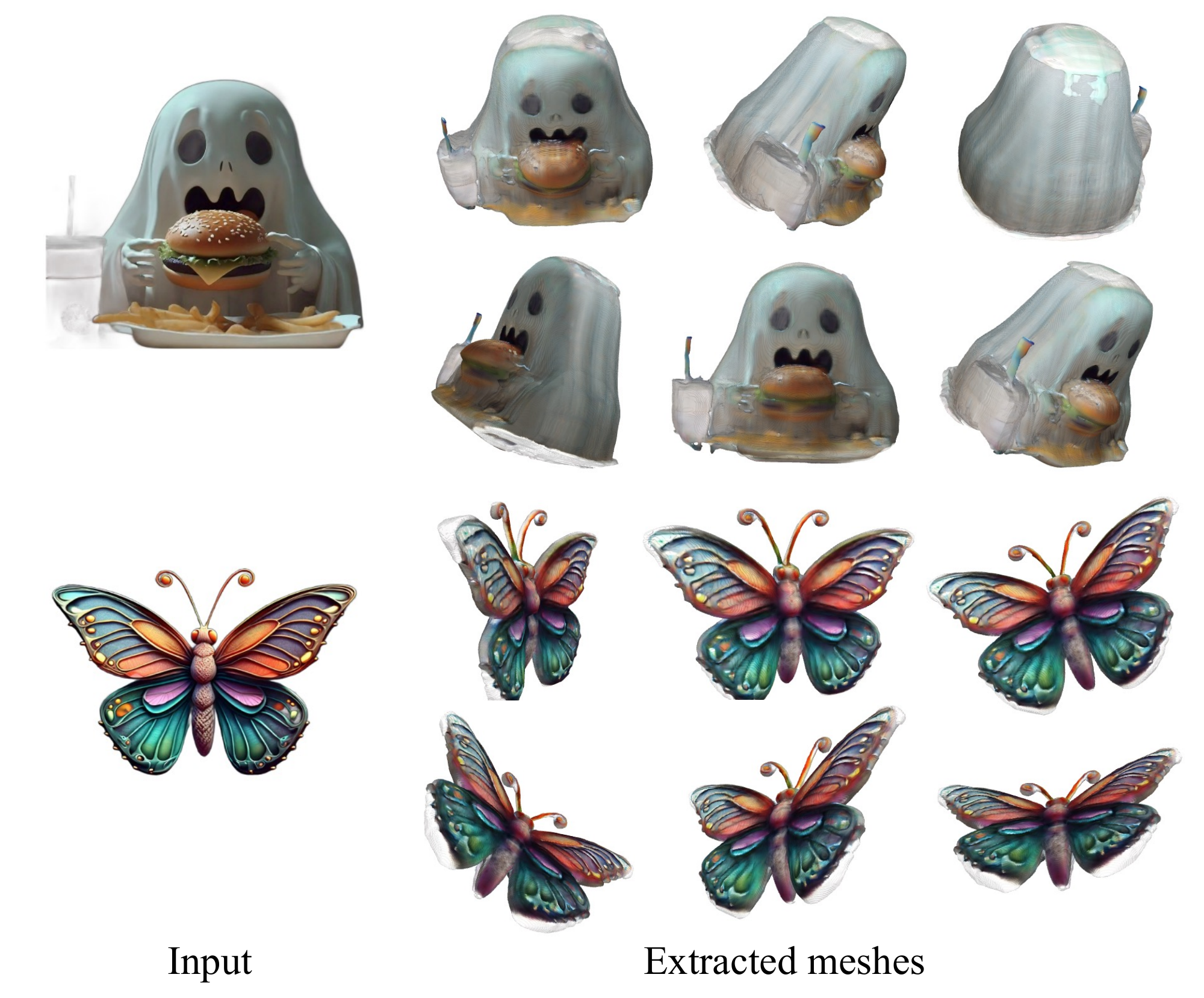}
\caption{\textbf{Samples of converted meshes}. We can create detailed and accurate meshes from the generated NeRF results in seconds.}
     \label{fig:mesh}
\end{figure}

We use the marching cubes algorithm~\cite{lorensen1998marching} to extract meshes from the generated NeRF results. Visualizations of sample converted meshes are shown in Figure~\ref{fig:mesh}.

\end{document}